\documentclass[10pt,a4paper,twoside]{article}

\usepackage[T1]{fontenc}
\usepackage[utf8]{inputenc}
\IfFileExists{newtxtext.sty}{%
  \usepackage{newtxtext}\usepackage{newtxmath}}{%
  \usepackage{mathptmx}}

\usepackage[a4paper,top=30mm,bottom=25mm,left=22mm,right=22mm,
            headheight=14pt,headsep=12mm,footskip=15mm]{geometry}

\usepackage{amsmath}
\usepackage{graphicx}
\usepackage{booktabs}
\usepackage{array}
\usepackage{fancyhdr}
\usepackage{titlesec}
\usepackage{multirow}
\usepackage[hidelinks]{hyperref}
\usepackage[font=normalsize,labelsep=colon,skip=6pt]{caption}
\usepackage{enumitem}
\usepackage{siunitx} 
\usepackage{cite} 
\usepackage[hidelinks]{hyperref}

\usepackage{algorithm}
\usepackage{algpseudocode}

\graphicspath{{figures/}}

\newcommand{\papertitle}{Domain-Aware Conformal Prediction for Trustworthy Spacecraft Pose Estimation under Sim-to-Real Shift}
\newcommand{\shorttitle}{Domain-aware conformal spacecraft pose estimation} 
\newcommand{\shortauthors}{N. E. Chenni, A. Rathinam, D. Aouada}           
\newcommand{\copyrightauthors}{Nidhal Eddine Chenni, Arunkumar Rathinam and Djamila Aouada} 

\fancypagestyle{firstpage}{%
  \fancyhf{}
  \fancyhead[R]{\small JOINT I-SAIRAS \& ISPARO CONFERENCE 2026}
  \fancyfoot[C]{\small Copyright \copyright\ 2026 by \copyrightauthors. Published by the i-SAIRAS association with permission.}
  }

\titleformat{\section}{\fontsize{12}{14}\selectfont\bfseries}{\thesection.}{0.4em}{}
\titlespacing*{\section}{9.5mm}{14pt}{14pt}
\titleformat{\subsection}{\fontsize{11}{13}\selectfont\bfseries}{\thesubsection}{0.5em}{}
\titlespacing*{\subsection}{0pt}{14pt}{14pt}

\newcommand{\notation}[1]{\textbf{#1}}

\renewenvironment{abstract}{%
  \par\vspace{12pt}
  \begin{list}{}{\setlength{\leftmargin}{9.5mm}\setlength{\rightmargin}{10.8mm}
                 \setlength{\topsep}{0pt}\setlength{\parsep}{0pt}}
  \item\relax{\fontsize{12}{14}\selectfont\bfseries Abstract}\par\vspace{2pt}}%
  {\end{list}\vspace{4pt}}

\makeatletter
\renewcommand\@biblabel[1]{[#1]}
\renewcommand\refname{\hspace*{9.5mm}References}
\makeatother

\begin{document}
\thispagestyle{firstpage}

\begin{center}
  \vspace*{8mm}
  {\fontsize{18}{21}\selectfont\bfseries \papertitle\par}
  \vspace{12mm}
  {\itshape
    Nidhal Eddine Chenni, Arunkumar Rathinam and Djamila Aouada\\[3pt]
    SnT, University of Luxembourg, Luxembourg\\[3pt]
    \{nidhal.chenni, arunkumar.rathinam, djamila.aouada\}@uni.lu\par}
  \vspace{8mm}
\end{center}

\begin{abstract}
Autonomous rendezvous and proximity operations require not only an accurate estimate of the pose of the target spacecraft, but also a reliable indication of when this estimate can be trusted. Conformal prediction provides such an indication with a formal guarantee, by producing pose uncertainty sets that contain the true pose with a specified probability. Such a guarantee, however, assumes that calibration and test data follow the same distribution, an assumption that is violated in orbital perception, where labelled calibration images are available only from synthetic renderings while operational images are real. In this work, we show on the SPEED+ dataset that conformal pose estimation calibrated on synthetic images loses its coverage guarantee on real images, and that many of the resulting failures are silent: the predicted pose set appears valid but excludes the true pose. To address this, we calibrate on synthetic images translated to the appearance of real images. Since translation alters appearance while preserving geometry, the synthetic pose labels remain valid and no labelled real images are required. Experiments with two translation models show that this calibration restores coverage above the specified level for miscoverage levels of 0.1 and above, and reduces silent failures from up to 43.5\% to at most 3.3\% of the test images, bringing conformal pose estimation a step closer to trustworthy deployment in space.
 
\vspace{4pt}
\end{abstract}

\section{Introduction}
\vspace{-8pt}
Estimating the relative position and orientation (6-DoF pose) of a target spacecraft with respect to a servicer is a key enabling technology for autonomous rendezvous and proximity operations, such as on-orbit servicing, inspection and active debris removal \cite{sharma2018pose,pauly2023survey}. When the target is non-cooperative, it provides no information about its own state, and its pose must therefore be estimated from onboard sensors. Monocular cameras are an attractive choice owing to their low mass, power consumption and cost \cite{sharma2018pose,pauly2023survey}, and methods based on deep neural networks have demonstrated high accuracy on monocular spacecraft pose estimation benchmarks \cite{chen2019satellite,park2023satellite}. Training these networks requires large datasets of images annotated with accurate pose labels, which are difficult and costly to acquire in orbit. Networks are therefore typically trained on synthetic images rendered from a 3D model of the target. However, these images do not fully capture the visual characteristics of real imagery, such as surface textures and complex illumination effects. As a result, networks trained exclusively on synthetic data suffer a significant drop in accuracy when evaluated on real images \cite{singh2025bridging}. This synthetic-to-real (sim2real) gap has become a major focus of the space community, and several datasets have been proposed to study it, including SPEED+ \cite{park2022speed+}, SPARK \cite{musallam2021spacecraft} and SPADES \cite{rathinam2024spades}. Numerous domain adaptation methods have since been proposed to reduce this gap, with encouraging results \cite{park2023satellite,park2024robust,chenni2026unsupervised}. Nevertheless, even the best-performing methods still produce erroneous estimates on a subset of real images, and aggregate metrics such as the mean error may conceal these failures \cite{chenni2026unsupervised}. 

An important question therefore remains open: how can the system determine whether an estimated pose can be trusted? This question is critical in practice, since an inaccurate pose that goes undetected may lead to unsafe manoeuvres of the servicer. Each estimate should therefore be accompanied by a reliable measure of its uncertainty, so that unreliable estimates can be identified and handled accordingly. The confidence scores produced by the network itself are not suitable for this purpose, as they are not calibrated and may be particularly misleading on real images, where errors are most likely.

Several approaches have been proposed to quantify the uncertainty of deep neural networks, including Monte Carlo dropout \cite{gal2016dropout}, deep ensembles \cite{lakshminarayanan2017simple} and the direct prediction of error variances, which has also been applied to spacecraft keypoint detection \cite{wang2024monocular}. While informative, these estimates come with no formal guarantee, and their quality has been shown to deteriorate under distribution shift \cite{ovadia2019can}. Conformal prediction \cite{angelopoulos2023conformal} provides an alternative with formal guarantees: using a held-out labelled calibration set, it converts the output of any pre-trained model into a prediction set that contains the ground truth with a specified probability. Because it requires neither retraining nor assumptions on the model, it can be applied to an already validated network, which makes it particularly well suited to space applications.

Yang and Pavone \cite{yang2023object} extended conformal prediction, previously applied mainly to classification and low-dimensional regression, to object pose estimation. Their method calibrates prediction regions around detected keypoints and propagates them to a set of poses that contains the ground-truth pose with the same probability, together with worst-case bounds on the estimation error. Wang et al. \cite{wang2025satellite} subsequently adapted this approach to satellite pose estimation on synthetic images. The validity of these guarantees, however, relies on the assumption that calibration and test data are exchangeable, which in practice requires them to follow the same distribution. Both works satisfy this assumption by calibrating and testing within the same domain. In spacecraft pose estimation, this is not possible: labelled calibration data are available only for synthetic images, whereas operational images are real. Since the network is less accurate on real images, prediction regions calibrated on synthetic data can be expected to be too narrow, and the true pose may fall outside the predicted set more often than specified. To the best of our knowledge, this effect has not yet been studied.

In this work, we first quantify this effect on SPEED+ and show that the coverage obtained on real images falls well below the specified level. We then address it by observing that recovering the specified coverage requires a calibration set that is both labelled and representative of real images. Since the pose label of an image is determined by its geometry rather than its appearance, a synthetic image whose appearance is modified while its geometry is preserved retains a valid label. Image-to-image translation models perform precisely this transformation, and have been used successfully to reduce the sim2real gap on SPEED+, for instance by the winning team of the Sunlamp category of the 2021 Satellite Pose Estimation Competition \cite{park2023satellite}. 
We therefore use translation not to train the network, but to construct a calibration set that approximates the real domain. The conformal method of \cite{yang2023object} is otherwise left unchanged, and no labelled real images are required.

The contributions of this paper are as follows:
\begin{itemize}
    \item We apply the conformal keypoint detection and pose uncertainty propagation of \cite{yang2023object} to spacecraft pose estimation on SPEED+, and analyse why its coverage guarantee no longer holds when calibration images are synthetic and test images are real.
    \item We quantify this loss of coverage on the real Lightbox and Sunlamp domains, and show that a large share of the uncovered images are silent failures, in which a non-empty pose set excludes the true pose.
    \item We propose to calibrate on synthetic images translated to the appearance of real images, and show that this restores coverage above the specified level for miscoverage levels of 0.1 and above and reduces silent failures from up to 43.5\% to at most 3.3\% of the test images, without requiring labelled real images.
\end{itemize}

The remainder of this paper is organized as follows. Section~\ref{sec:related} reviews related work on spacecraft pose estimation under the sim2real gap, uncertainty quantification and image translation. Section~\ref{sec:conformal} presents the conformal pose estimation method and analyzes why its guarantee is lost under the sim2real gap. Section~\ref{sec:method} introduces calibration with translated images. Section~\ref{sec:experiments} reports the experimental results on SPEED+, Section~\ref{sec:discussion} discusses the limitations of the proposed approach, and Section~\ref{sec:conclusion} provides summary and concludes the paper.
\section{Related Works}
\label{sec:related}
\vspace{-8pt}
\notation{Spacecraft pose estimation under the sim2real gap.} Deep learning methods for monocular spacecraft pose estimation fall into two families \cite{pauly2023survey}: Direct approaches that regress the pose directly from the image, and Hybrid approaches that first predict the 2D locations of predefined keypoints and then recover the pose with a Perspective-n-Point (PnP) solver \cite{lepetit2009ep}. The latter have generally achieved higher accuracy, as shown by the results of the SPEED challenge \cite{kisantal2020satellite,chen2019satellite}, and form the basis of the present work. As these methods matured on synthetic data, attention shifted to their performance on real images. The 2021 Satellite Pose Estimation Competition, built on SPEED+ \cite{park2022speed+,park2023satellite}, made the sim2real gap its central challenge, and the top-performing entries addressed it through strategies such as extensive data augmentation, image translation and self-training on unlabelled real images. Building on these results, later works further reduced the gap through multi-task learning with online refinement \cite{park2024robust}, unsupervised domain adaptation \cite{perez2022spacecraft,chenni2026unsupervised} and, when a small number of labelled real images is available, supervised adaptation \cite{singh2025bridging}. Throughout this line of work, progress has been measured by average accuracy, while the reliability of individual estimates has received comparatively little attention.

\notation{Uncertainty quantification and conformal prediction for pose.} Assessing the reliability of an estimate requires quantifying its uncertainty. For deep networks, this is commonly done with Bayesian approximations such as Monte Carlo dropout \cite{gal2016dropout}, with deep ensembles \cite{lakshminarayanan2017simple}, or by training the network to predict its own error variance. The latter approach has been applied to spacecraft pose estimation by Wang et al. \cite{wang2024monocular}, who used the predicted keypoint uncertainty to improve pose accuracy. Such estimates are informative, but they offer no guarantee that the true value lies within the predicted range. Conformal prediction \cite{vovk2005algorithmic,angelopoulos2023conformal} provides such a guarantee: given a held-out labelled calibration set, it constructs prediction sets that contain the ground truth with a specified probability, provided that calibration and test data are exchangeable. Yang and Pavone \cite{yang2023object} extended this framework to object pose estimation by calibrating prediction regions around keypoints detected from heatmaps and propagating them to a pose uncertainty set with worst-case error bounds. Wang et al. \cite{wang2025satellite} later adapted this approach to satellite pose estimation, replacing the heatmaps with a regression network that predicts keypoint uncertainty and calibrating each keypoint separately. In both works, however, calibration and test images are drawn from the same domain, and Wang et al. identify the extension to real imagery as future work. Applying these methods to spacecraft therefore requires a labelled calibration set that is representative of real images, which is precisely what is unavailable in practice. Section~\ref{sec:breaks} analyses the consequences of this mismatch.

\notation{Image translation for the sim2real gap.} A natural way to obtain such a calibration set is to change the appearance of labelled synthetic images while preserving their content. Unpaired image-to-image translation models, introduced by CycleGAN \cite{zhu2017unpaired}, learn this mapping between two domains without requiring paired examples. In domain adaptation, they are commonly used to render labelled source images in the style of the target domain so that the source labels can be reused for training \cite{hoffman2018cycada}, and later models such as DUNIT \cite{bhattacharjee2020dunit} better preserve object content by translating object instances separately from the background. In spacecraft pose estimation, synthetic-to-real translation has likewise been used to reduce the sim2real gap on SPEED+ \cite{wang2023bridging}. In our work, translation serves a different purpose: rather than improving the network, it is used to construct a calibration set that is representative of real images, as described in Section~\ref{sec:method}. We consider both CycleGAN and DUNIT to assess how the choice of translation model affects the recovered coverage.
\section{Conformal Pose Estimation}
\label{sec:conformal}
\vspace{-8pt}
This section reviews the method of Yang and Pavone~\cite{yang2023object}, which we apply to spacecraft pose estimation. The method proceeds in two stages. Conformal keypoint detection first converts the output of a keypoint network into prediction regions that contain the true keypoints with a specified probability. Geometric uncertainty propagation then maps these regions to a set of poses that contains the true pose with the same probability, from which a pose estimate is derived.

\subsection{Problem setup}
\vspace{-8pt}
The target spacecraft is described by $K$ keypoints with known 3D coordinates $Y_1, \dots, Y_K$ in its body frame ($K = 11$ for SPEED+). For an image $x$ acquired under the true pose $(R, t) \in \mathrm{SE}(3)$, the true 2D keypoints are given by
\begin{equation}
    y_k = \pi(R Y_k + t), \qquad k = 1, \dots, K,
    \label{eq:projection}
\end{equation}
where $\pi$ denotes the perspective projection of the calibrated camera. A keypoint network $f$ predicts one heatmap per keypoint. Following~\cite{yang2023object}, each heatmap is normalized to a probability distribution over pixels, from which we read the most likely location $q_k$ and its probability $p_k$. In a conventional pipeline, the pose is then obtained by solving the Perspective-n-Point (PnP) problem on the correspondences $\{q_k \leftrightarrow Y_k\}$~\cite{lepetit2009ep}, which gives a single estimate with no indication of its reliability.

\subsection{Inductive conformal prediction}
\label{sec:icp}
\vspace{-8pt}
Inductive conformal prediction~\cite{vovk2005algorithmic,angelopoulos2023conformal} turns the output of any trained model into a prediction set with a coverage guarantee. Its central ingredient is a non-conformity score $s(x, y)$, which measures how poorly a label $y$ agrees with the prediction of the model on input $x$. Given $n$ labelled calibration samples $(x_i, y_i)$, disjoint from the training data, the scores $s_i = s(x_i, y_i)$ are computed and, for a miscoverage level $\alpha \in [0, 1]$, the threshold $\tau$ is set to the $\lceil (n+1)(1-\alpha) \rceil$-th smallest score. The prediction set for a new input $x$ then contains every label whose score does not exceed this threshold:
\begin{equation}
    C(x) = \left\{ y : s(x, y) \le \tau \right\}.
    \label{eq:icpset}
\end{equation}
If the calibration samples and the test sample $(x, y)$ are exchangeable, that is, if their joint distribution is invariant to permutations, then
\begin{equation}
    1 - \alpha \;\le\; \mathbb{P}\left( y \in C(x) \right) \;\le\; 1 - \alpha + \frac{1}{n+1},
    \label{eq:coverage}
\end{equation}
where the upper bound holds when scores have no ties~\cite{vovk2005algorithmic}. The argument rests on ranks: under exchangeability, the score of the test sample is equally likely to occupy any of the $n + 1$ positions among all scores, and therefore falls below $\tau$ with probability at least $1 - \alpha$. The guarantee is marginal, i.e., it holds on average over the draw of calibration and test samples, and it makes no assumption on the model or on the distribution of its errors.

\subsection{Conformal keypoint detection}
\label{sec:ckd}
\vspace{-8pt}
To apply this framework to keypoints, Yang and Pavone define the score of an image as
\begin{equation}
    s(x, y) = \max_{k = 1, \dots, K} \; p_k \, \lVert y_k - q_k \rVert .
    \label{eq:score}
\end{equation}
Each term weights the localisation error of a keypoint by the confidence of the network, so that the score is largest when the network is confidently wrong. Taking the maximum over keypoints makes the score depend on the worst keypoint in the image. Consequently, the condition $s(x, y) \le \tau$ in Eq.~\eqref{eq:icpset} holds if and only if every keypoint satisfies $p_k \lVert y_k - q_k \rVert \le \tau$, and the prediction set decomposes into one disk per keypoint:
\begin{equation}
    C_k(x) = \left\{ y \in \mathbb{R}^2 : \lVert y - q_k \rVert \le \frac{\tau}{p_k} \right\}, \qquad k = 1, \dots, K.
    \label{eq:disk}
\end{equation}
Each disk is centred at the predicted keypoint, and its radius is inversely proportional to the confidence of the network and proportional to the threshold $\tau$, which reflects the errors observed on the calibration set. By Eq.~\eqref{eq:coverage}, the $K$ true keypoints lie simultaneously inside their disks with probability at least $1 - \alpha$. This joint coverage, rather than a per-keypoint one, is what allows the guarantee to be transferred to the pose.

\subsection{From keypoint regions to a pose uncertainty set}
\label{sec:purse}
\vspace{-8pt}
The keypoint disks constrain the pose through the projection in Eq.~\eqref{eq:projection}. Yang and Pavone define the pose uncertainty set as the set of all poses whose projected keypoints fall inside their respective disks:
\begin{equation}
    S(x) = \left\{ (R, t) \in \mathrm{SE}(3) : \pi(R Y_k + t) \in C_k(x), \;\; k = 1, \dots, K \right\}.
    \label{eq:poseset}
\end{equation}
Since the true keypoints are the projections of the 3D keypoints under the true pose, they lie in all disks if and only if the true pose belongs to $S(x)$. The pose set therefore inherits the guarantee of the keypoint disks,
\begin{equation}
    \mathbb{P}\left( (R, t) \in S(x) \right) \ge 1 - \alpha,
    \label{eq:posecoverage}
\end{equation}
without any assumption on the pose solver. This equivalence also provides a direct way to evaluate pose coverage on labelled data, by checking whether the ground-truth keypoints lie in their disks.

The set $S(x)$ admits an explicit algebraic description. Writing the pose as the vector $s = [\mathrm{vec}(R)^\top, t^\top]^\top \in \mathbb{R}^{12}$, the camera-frame keypoint $R Y_k + t$ is linear in $s$. Multiplying each disk constraint by the squared depth of the keypoint, which is positive for points in front of the camera, turns it into a quadratic inequality $s^\top A_k s \le 0$, while the requirement of positive depth becomes a linear inequality $b_k^\top s > 0$. Here $A_k$ and $b_k$ depend only on the disk, the 3D keypoint and the camera intrinsics~\cite{yang2023object}. $S(x)$ is thus the intersection of $\mathrm{SE}(3)$ with $K$ quadratic and $K$ linear constraints. It is in general nonconvex, and it may be empty when the disks are mutually inconsistent. Since an empty set cannot contain the true pose, this event occurs with probability at most $\alpha$ and itself indicates an unreliable prediction.

\subsection{Average pose}
\label{sec:ransag}
\vspace{-8pt}
The pose set characterizes the uncertainty, but does not directly provide a point estimate. Sampling $S(x)$ directly is difficult because of its nonconvex constraints, whereas sampling the keypoint disks is straightforward. Yang and Pavone therefore propose random sample averaging (RANSAG). At each of $T$ iterations, three distinct keypoints are selected at random, a point is sampled within each of their disks, and the resulting minimal problem is solved with P3P. The candidate poses that satisfy all constraints of $S(x)$ are retained. If no candidate is retained, poses are instead obtained by solving PnP on points sampled in all $K$ disks. The average pose $(\bar R, \bar t)$ is computed from the retained samples, by projecting the arithmetic mean of the rotations onto $\mathrm{SO}(3)$ and averaging the translations. Yang and Pavone~\cite{yang2023object} further derive worst-case bounds on the error of the average pose through semidefinite relaxations, which are not considered in this work. The pose set thus accompanies the average pose with an uncertainty statement that holds with a specified probability, and an empty pose set indicates that the estimate should not be trusted.
\subsection{Exchangeability under the sim2real gap}
\label{sec:breaks}
\vspace{-8pt}
In spacecraft pose estimation, reliable pose labels are available only for synthetic images. The calibration set is therefore necessarily synthetic, whereas the images acquired during operations are real. Calibration and test scores then follow two different distributions, and the rank argument of Section~\ref{sec:icp} no longer applies.

The direction of the resulting error can be anticipated. Owing to the sim2real gap, the keypoint network makes larger errors on real images than on synthetic ones. If its confidence $p_k$ does not decrease accordingly, the network becomes confidently wrong on real images, which is precisely the situation penalized by the score in Eq.~\eqref{eq:score}. Real scores are therefore expected to be shifted towards larger values than synthetic ones. The threshold $\tau_{\text{syn}}$ estimated on synthetic data is then too small for real images, the disks in Eq.~\eqref{eq:disk} are too narrow, and the true keypoints, and hence the true pose, fall outside the predicted sets more often than specified. The coverage actually obtained on real images is the fraction of real scores below the synthetic threshold,
\begin{equation}
    \mathbb{P}_{x \sim \text{real}}\left( s(x, y) \le \tau_{\text{syn}} \right),
    \label{eq:realcov}
\end{equation}
which may fall well below the specified level $1 - \alpha$. This failure is silent: the prediction sets retain a plausible size and shape, and nothing in the output indicates that the guarantee no longer holds. Section~\ref{sec:experiments} quantifies this effect on SPEED+.

Recalibrating on labelled real images would restore exchangeability, but such images are not available in orbit. What is required instead is a calibration set that carries reliable labels while being representative of real images. Section~\ref{sec:method} describes how such a set can be constructed from synthetic data.

\section{Domain-Aware Calibration with Translated Images}
\label{sec:method}
\begin{figure}[t]
    \centering
    \includegraphics[width=\linewidth]{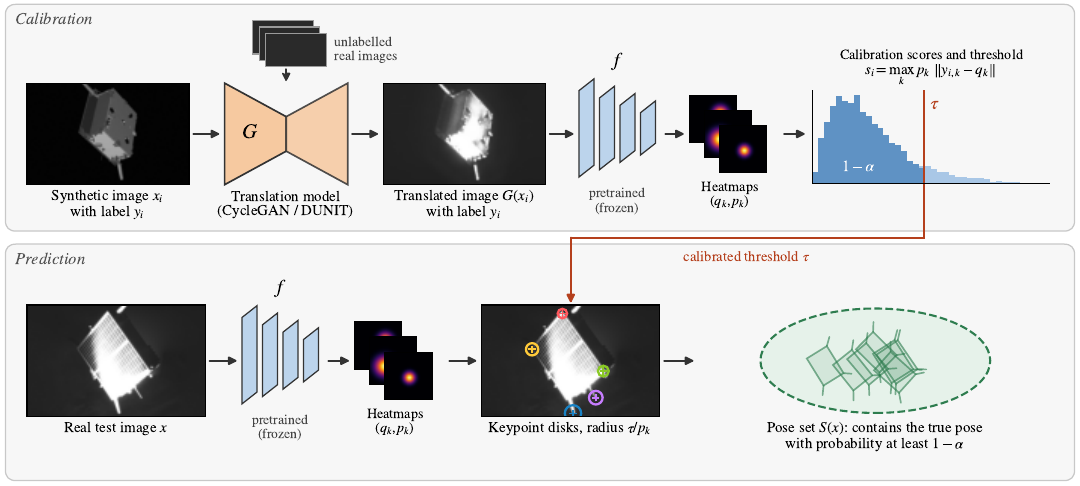}
    \caption{Overview of the proposed pipeline. Labelled synthetic calibration images are translated to the appearance of the real domain, and the nonconformity scores computed on the translated images by the pretrained keypoint network determine the threshold $\tau$. For each real test image, $\tau$ defines the keypoint disks, which are propagated to a pose set that contains the true pose with probability at least $1-\alpha$.}
    \label{fig:pipeline}
\end{figure}

\vspace{-8pt}
Section~\ref{sec:breaks} showed that the coverage guarantee of conformal pose estimation depends on the calibration set being representative of the test images, a condition that a synthetic calibration set does not satisfy for real images. We address this mismatch by modifying only the calibration set: each labelled synthetic calibration image is rendered in the appearance of the real domain by an image-to-image translation model $G$, and the nonconformity scores are computed on the translated images. The keypoint network, the score, the keypoint disks and the propagation to the pose described in Section~\ref{sec:conformal} remain unchanged. Figure~\ref{fig:pipeline} illustrates the resulting pipeline.

This construction relies on the fact that the pose label of an image is determined by the geometry of the spacecraft in the image, namely its position, orientation and silhouette, and not by its appearance. If translation modifies textures and illumination while preserving this geometry, the translated image $G(x_i)$ depicts the spacecraft under the same pose as $x_i$, and the pair $(G(x_i), y_i)$ is correctly labelled without any annotation of real images. The translation model should therefore alter appearance as much as needed to resemble the real domain, while retaining the structure of the spacecraft. We consider two unpaired translation models that pursue this objective in different ways. CycleGAN~\cite{zhu2017unpaired} learns a mapping between the two domains at the image level, constrained by a cycle-consistency loss that encourages the translated image to retain the content of its input. DUNIT~\cite{bhattacharjee2020dunit} additionally detects object instances and translates them separately from the background, which is expected to better preserve the structure of the spacecraft. Since the two real domains of SPEED+ differ substantially in illumination, a separate translation model is trained for each of them, from synthetic to Lightbox and from synthetic to Sunlamp, using synthetic images and unlabelled real images of the corresponding domain. The keypoint network is not retrained.

Algorithm~\ref{alg:method} summarises the procedure. During calibration, the synthetic calibration images are translated, and the score of Eq.~\eqref{eq:score} is computed on each translated image using the original synthetic label. The threshold $\tau$ is then obtained as in Section~\ref{sec:icp}. At test time, each real image is processed exactly as in Section~\ref{sec:conformal}, with this threshold in place of the one estimated on synthetic images.

\begin{algorithm}[t]
\caption{Domain-aware conformal pose estimation}
\label{alg:method}
\begin{algorithmic}[1]
\Require keypoint network $f$; labelled synthetic calibration set $\{(x_i, y_i)\}_{i=1}^{n}$; unlabelled real images $\mathcal{U}$ of the target domain; miscoverage level $\alpha \in [0, 1]$
\Statex \textit{Calibration}
\State Train the translation model $G$ on synthetic images and $\mathcal{U}$
\For{$i = 1, \dots, n$}
    \State $\tilde{x}_i \gets G(x_i)$ \Comment{appearance translated, label $y_i$ kept}
    \State Obtain $(q_k, p_k)_{k=1}^{K}$ from the heatmaps of $f(\tilde{x}_i)$
    \State $s_i \gets \max_{k} \, p_k \lVert y_{i,k} - q_k \rVert$ \Comment{Eq.~\eqref{eq:score}}
\EndFor
\State $\tau \gets \lceil (n+1)(1-\alpha) \rceil$-th smallest value of $\{s_i\}_{i=1}^{n}$
\Statex \textit{Prediction for a real test image $x$}
\State Obtain $(q_k, p_k)_{k=1}^{K}$ from the heatmaps of $f(x)$
\State $C_k(x) \gets \{ y \in \mathbb{R}^2 : \lVert y - q_k \rVert \le \tau / p_k \}$ for $k = 1, \dots, K$ \Comment{Eq.~\eqref{eq:disk}}
\State Define the pose set $S(x)$ from $\{C_k(x)\}_{k=1}^{K}$ \Comment{Eq.~\eqref{eq:poseset}}
\State Compute the average pose $(\bar R, \bar t)$ with RANSAG 
\State \Return $S(x)$, $(\bar R, \bar t)$
\end{algorithmic}
\end{algorithm}

If translated images followed exactly the distribution of real images, calibration and test samples would be exchangeable, and the guarantee of Eq.~\eqref{eq:coverage} would hold again. In practice, translation is only approximate, and the remaining mismatch can be related to the loss of coverage. Let $P$ and $Q$ denote the distributions of the scores on translated and real images, respectively. The coverage obtained on real images then satisfies~\cite{barber2023conformal}
\begin{equation}
    \mathbb{P}\left( s(x, y) \le \tau \right) \ge 1 - \alpha - d_{TV}(P, Q),
    \label{eq:tv}
\end{equation}
where $d_{TV}$ denotes the total variation distance. The loss of coverage is thus controlled by the discrepancy between score distributions, rather than by the visual discrepancy between images. Translated images therefore need not be indistinguishable from real ones; it suffices that the keypoint network errs on them in a similar way. Since $Q$ is unknown in practice, Eq.~\eqref{eq:tv} does not provide a guarantee at test time, but it clarifies the condition under which calibration on translated images can be expected to recover the specified coverage.
\section{Experiments}
\label{sec:experiments}
\vspace{-8pt}
The experiments address three questions. First, does conformal pose estimation calibrated on synthetic images retain its coverage on real images? Second, does calibration on translated images recover the specified coverage? Third, how does the choice of calibration set affect the accuracy of the average pose?

\subsection{Experimental setup}
\vspace{-4pt}
\notation{Dataset.} All experiments are conducted on SPEED+~\cite{park2022speed+}. The synthetic domain contains 59{,}960 rendered images of the Tango spacecraft, split into 47{,}966 training and 11{,}994 validation images. The two real domains, Lightbox (6{,}740 images) and Sunlamp (2{,}791 images), were acquired with a physical mockup of Tango under two illumination conditions: Lightbox reproduces diffuse Earth albedo, whereas Sunlamp uses a direct light source that produces strong specular reflections and lens flare. All real images are used as test images.

\notation{Implementation.} The keypoint network is a heatmap-based model with an EfficientNet-B5 backbone~\cite{tan2019efficientnet}, trained on the synthetic training images only. The calibration set is drawn from the synthetic validation images. For each real domain, a CycleGAN and a DUNIT model are trained at a resolution of $640 \times 400$ on the synthetic images and the real images of that domain, the latter being used without their pose labels. The synthetic images used to train the translation models include the calibration images.

\notation{Calibration protocols.} All protocols use the same keypoint network and the same test images, and differ only in the calibration set. In \emph{Synthetic}, the scores are computed on the synthetic calibration images, which is the setting available in practice without translation. In \emph{CycleGAN} and \emph{DUNIT}, the calibration images are first translated to the corresponding real domain.

\notation{Metrics.} The \emph{coverage} is the fraction of test images whose ground-truth pose lies in the pose set $S(x)$; a valid calibration achieves a coverage of at least $1 - \alpha$. Each test image that is not covered falls into one of two cases. Either the pose set is \emph{empty}, in which case the estimate is explicitly flagged as unreliable, or the pose set is non-empty but does not contain the ground truth, in which case the failure is \emph{missed}. The latter case is the most critical for operations, since nothing in the output indicates that the estimate is wrong. The accuracy of the average pose is evaluated by the median rotation error, the median translation error and the pose score $E$ used in the Satellite Pose Estimation Competition~\cite{park2023satellite}, which sums the mean rotation error in radians and the mean normalised translation error.

\subsection{Coverage on real images}

\begin{figure}[t]
    \centering
    \includegraphics[width=0.9\linewidth]{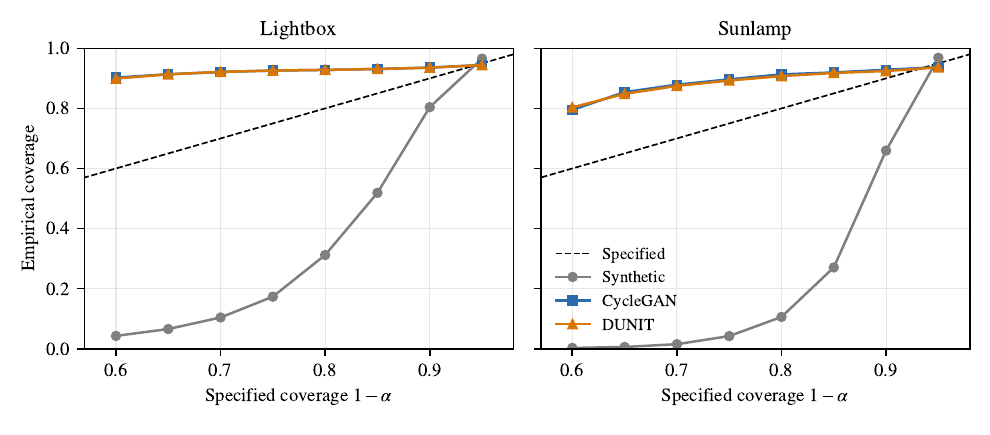}
    \caption{Empirical versus specified coverage on real test images for each calibration set. Points below the dashed diagonal indicate under-coverage.}
    \label{fig:coverage_curve}
    \vspace{-0.3cm}
\end{figure}

Table~\ref{tab:coverage} reports the coverage obtained under each protocol for $\alpha \in \{0.1, 0.2, 0.3\}$, and Figure~\ref{fig:coverage_curve} extends the comparison to $\alpha \in [0.05, 0.4]$. With synthetic calibration, the coverage on real images falls well below the specified level and degrades rapidly as $\alpha$ increases. At $\alpha = 0.1$, it reaches 80.4\% on Lightbox and 66.0\% on Sunlamp instead of the specified 90\%, and at $\alpha = 0.3$ it drops to 10.4\% and 1.5\%, respectively. The loss is larger on Sunlamp, consistent with its stronger departure from the synthetic illumination. At $\alpha = 0.05$, synthetic calibration does reach the specified coverage (96.6\% and 96.8\%), which indicates that the synthetic threshold exceeds the scores of most real images only for very small values of $\alpha$.

Calibration on translated images restores valid coverage for all $\alpha \ge 0.1$ on both domains. At $\alpha = 0.1$, CycleGAN and DUNIT reach 93.6\% and 93.5\% on Lightbox, and 92.9\% and 92.4\% on Sunlamp. At $\alpha = 0.05$, the coverage remains slightly below the specified 95\% (94.5\% and 93.7\% with CycleGAN). The coverage obtained with translated calibration, however, varies little with $\alpha$: on Lightbox, it decreases only from 93.6\% to 92.1\% when $\alpha$ increases from 0.1 to 0.3. The calibration is therefore conservative for larger values of $\alpha$. This behaviour suggests that the scores of most real images lie well below the thresholds obtained from translated images, while a small fraction of real images, on which the network fails severely, have scores that exceed all thresholds considered.

\begin{table}[t]
\centering
\caption{Coverage, empty-set rate and missed-failure rate (\%) on real test images. A missed failure is a non-empty pose set that does not contain the ground-truth pose. The specified coverage is $1 - \alpha$.}
\label{tab:coverage}
\small
\setlength{\tabcolsep}{4pt}
\begin{tabular}{llccccccccc}
\toprule
 & & \multicolumn{3}{c}{$\alpha = 0.1$} & \multicolumn{3}{c}{$\alpha = 0.2$} & \multicolumn{3}{c}{$\alpha = 0.3$} \\
\cmidrule(lr){3-5} \cmidrule(lr){6-8} \cmidrule(lr){9-11}
Domain & Calibration & Cov.$\uparrow$ & Empty & Missed$\downarrow$ & Cov.$\uparrow$ & Empty & Missed$\downarrow$ & Cov.$\uparrow$ & Empty & Missed$\downarrow$ \\
\midrule
\multirow{3}{*}{Lightbox} & Synthetic & 80.4 & 11.6 & 8.0 & 31.2 & 27.5 & 41.3 & 10.4 & 46.1 & 43.5 \\
 & CycleGAN & 93.6 & 3.5 & 2.9 & 92.8 & 5.0 & 2.2 & 92.1 & 6.5 & 1.3 \\
 & DUNIT & 93.5 & 3.6 & 2.9 & 92.8 & 5.0 & 2.2 & 92.2 & 6.5 & 1.4 \\
\midrule
\multirow{3}{*}{Sunlamp} & Synthetic & 66.0 & 15.9 & 18.1 & 10.6 & 48.5 & 40.9 & 1.5 & 72.4 & 26.0 \\
 & CycleGAN & 92.9 & 4.2 & 2.9 & 91.3 & 6.1 & 2.6 & 87.9 & 9.0 & 3.1 \\
 & DUNIT & 92.4 & 4.6 & 3.0 & 90.8 & 6.5 & 2.8 & 87.5 & 9.2 & 3.3 \\
\bottomrule
\end{tabular}
\vspace{-0.5cm}
\end{table}

\begin{figure}[t]
    \centering
    \includegraphics[width=0.9\linewidth]{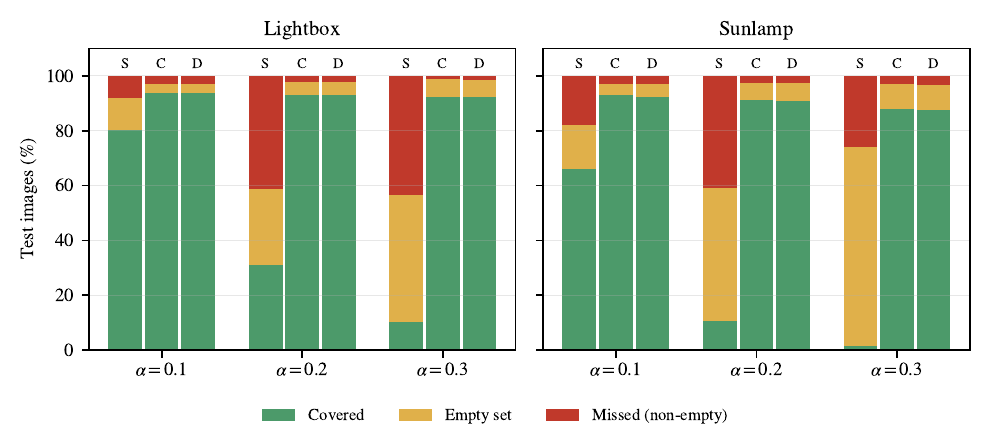}
    \caption{Outcome of conformal pose estimation on real test images for synthetic (S), CycleGAN (C) and DUNIT (D) calibration: pose set containing the ground truth (covered), empty pose set, and non-empty pose set excluding the ground truth (missed).}
    \label{fig:outcomes}
\end{figure}

The decomposition of the uncovered images in Table~\ref{tab:coverage} and Figure~\ref{fig:outcomes} shows that the consequences of synthetic calibration are more severe than the coverage alone suggests. With synthetic calibration, a large part of the uncovered images are missed failures: at $\alpha = 0.1$, 8.0\% of the Lightbox images and 18.1\% of the Sunlamp images receive a non-empty pose set that excludes the true pose, and this proportion exceeds 40\% on both domains at $\alpha = 0.2$. Empty sets also become frequent, reaching 72.4\% of the Sunlamp images at $\alpha = 0.3$, so that most estimates would be rejected. With translated calibration, the missed-failure rate remains between 1.3\% and 3.3\% on both domains and for all values of $\alpha$, and most uncovered images are instead flagged by an empty pose set.

\begin{table}[t]
\centering
\caption{Accuracy of the average pose on real test images: median rotation error (deg), median translation error (m) and pose score $E$.}
\label{tab:accuracy}
\small
\setlength{\tabcolsep}{4pt}
\begin{tabular}{llccccccccc}
\toprule
 & & \multicolumn{3}{c}{$\alpha = 0.1$} & \multicolumn{3}{c}{$\alpha = 0.2$} & \multicolumn{3}{c}{$\alpha = 0.3$} \\
\cmidrule(lr){3-5} \cmidrule(lr){6-8} \cmidrule(lr){9-11}
Domain & Calibration & Rot. & Trans. & $E$ & Rot. & Trans. & $E$ & Rot. & Trans. & $E$ \\
\midrule
\multirow{3}{*}{Lightbox} & Synthetic & 0.85 & 0.026 & 0.078 & 0.89 & 0.027 & 0.079 & 0.95 & 0.028 & 0.081 \\
 & CycleGAN & 2.64 & 0.118 & 0.152 & 1.30 & 0.047 & 0.093 & 1.01 & 0.033 & 0.082 \\
 & DUNIT & 2.50 & 0.111 & 0.146 & 1.29 & 0.047 & 0.093 & 1.02 & 0.034 & 0.083 \\
\midrule
\multirow{3}{*}{Sunlamp} & Synthetic & 1.35 & 0.042 & 0.098 & 1.44 & 0.045 & 0.101 & 1.54 & 0.048 & 0.103 \\
 & CycleGAN & 2.61 & 0.118 & 0.146 & 1.64 & 0.064 & 0.107 & 1.41 & 0.049 & 0.100 \\
 & DUNIT & 2.29 & 0.105 & 0.134 & 1.56 & 0.062 & 0.105 & 1.39 & 0.048 & 0.100 \\
\bottomrule
\end{tabular}
\end{table}

\subsection{Accuracy of the average pose}
\label{sec:exp_accuracy}

Since the average pose is computed from samples drawn within the keypoint disks, its accuracy depends on the size of the disks, and thus on the calibration set. Table~\ref{tab:accuracy} reports the accuracy of the average pose under each protocol. Synthetic calibration yields the most accurate average pose for $\alpha = 0.1$ and $\alpha = 0.2$, since its threshold, and therefore its disks, are smaller. This accuracy, however, comes with the loss of coverage reported above. With translated calibration, the larger disks increase the error of the average pose at $\alpha = 0.1$, with rotation errors between 2.3 and 2.6 degrees on both domains. This error decreases as $\alpha$ increases, and at $\alpha = 0.3$ translated calibration reaches an accuracy comparable to synthetic calibration on Lightbox (pose score 0.082 versus 0.081) and slightly better on Sunlamp (0.100 versus 0.103), while maintaining a coverage of approximately 92\% and 88\%, respectively. These results indicate a trade-off between coverage and the accuracy of the average pose, which, given the conservative behaviour of translated calibration, favours moderate values of $\alpha$.

\subsection{Comparison of translation models}

CycleGAN and DUNIT lead to nearly identical coverage, with differences of at most 0.5 percentage points in Table~\ref{tab:coverage}. DUNIT yields a slightly more accurate average pose at $\alpha = 0.1$, in particular on Sunlamp, where the median rotation error decreases from 2.61 to 2.29 degrees and the pose score from 0.146 to 0.134. These differences remain small compared with the gap between synthetic and translated calibration, which indicates that the recovery of coverage stems primarily from translating the calibration images rather than from the specific choice of translation model.

\subsection{Qualitative results}

\begin{figure}[t]
    \centering
    \includegraphics[width=\linewidth]{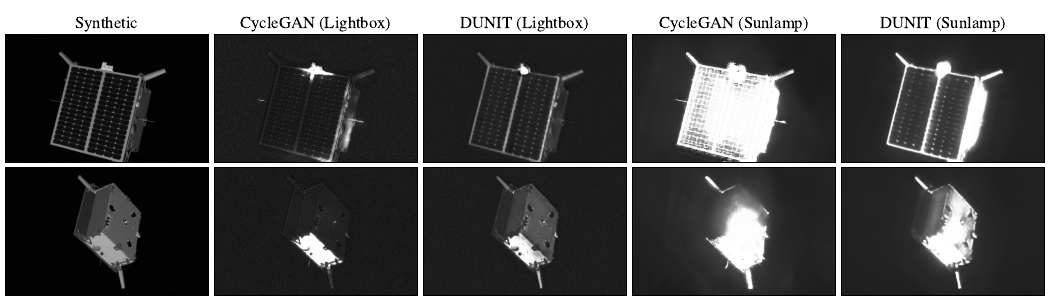}
    \caption{Two synthetic calibration images (left) and their translations to the Lightbox and Sunlamp domains by CycleGAN and DUNIT.}
    \label{fig:translation}
    \vspace{-0.2cm}
\end{figure}

\begin{figure}[h!]
    \centering
    \includegraphics[width=\linewidth]{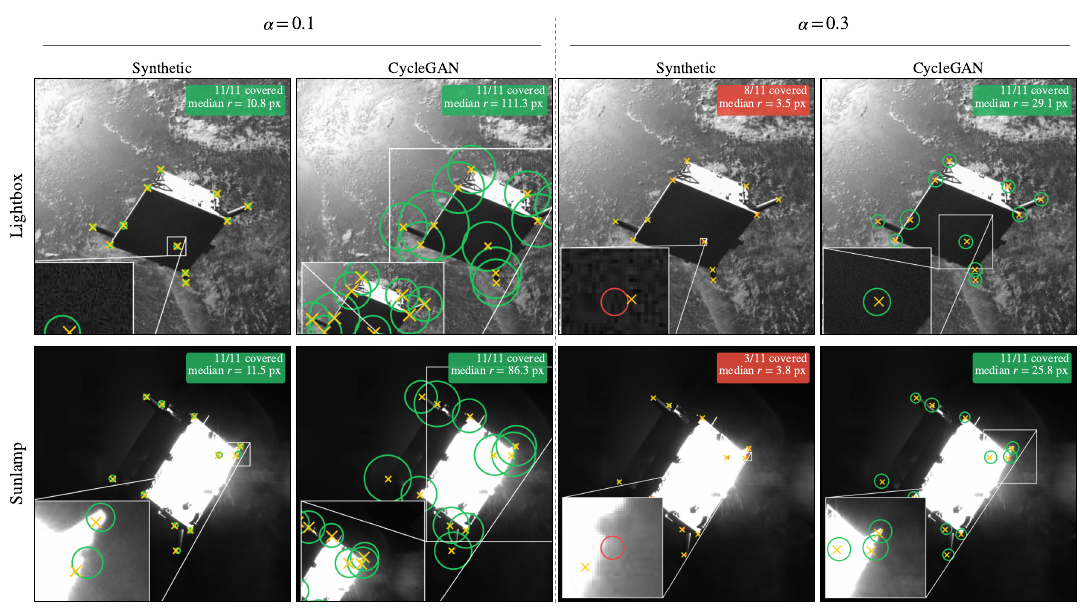}
\caption{Keypoint disks on a real Lightbox (top) and Sunlamp (bottom) test image at $\alpha = 0.1$ (left) and $\alpha = 0.3$ (right), obtained with synthetic calibration and with CycleGAN calibration. Yellow crosses indicate the ground-truth keypoints; green and red disks indicate covered and missed keypoints, respectively. The boxes in the bottom-left corner of each panel show a zoomed view of the region marked in the image; at $\alpha = 0.3$, they show a keypoint missed under synthetic calibration and the same region under translated calibration.}
    \label{fig:disks}
\end{figure}

Figure~\ref{fig:translation} shows two synthetic calibration images translated to each real domain. In all cases, the silhouette of the spacecraft and the position of its antennas are preserved, so that the synthetic keypoint labels remain valid for the translated images. The Lightbox translations introduce sensor noise, darker surfaces and localised specular highlights. The Sunlamp translations reproduce the strong overexposure and glare characteristic of this domain. CycleGAN tends to saturate large parts of the spacecraft body and to add diffuse glare to the background, whereas DUNIT confines the saturation to smaller regions and better preserves surface structures such as the pattern of the solar panel. This difference is consistent with the instance-level translation performed by DUNIT and with its slightly more accurate average pose on Sunlamp. The saturation introduced by the Sunlamp translations also makes some keypoints harder to localise on the calibration images, which may contribute to the conservative behaviour of translated calibration observed in Figure~\ref{fig:coverage_curve}. Figure~\ref{fig:disks} shows keypoint disks predicted on real test images of both domains. At $\alpha = 0.3$, synthetic calibration yields a median disk radius below 4 pixels, and several ground-truth keypoints fall outside their disks, whereas CycleGAN calibration increases the radius to between 26 and 29 pixels and covers all 11 keypoints. At $\alpha = 0.1$, both calibrations cover all keypoints in the examples shown, but the disks obtained with CycleGAN calibration are considerably larger, with a median radius between 86 and 111 pixels against about 11 pixels, which is consistent with the lower accuracy of the average pose reported in Section~\ref{sec:exp_accuracy}. The disks also vary in size within an image, being larger for keypoints on which the network is less confident.

\section{Discussion and Limitations}
\label{sec:discussion}

\notation{Silent failures under synthetic calibration.} The experiments confirm that the coverage guarantee of conformal pose estimation does not transfer from synthetic to real images. The most concerning aspect of this failure is not the loss of coverage itself, but the fact that a large share of the uncovered images receive a pose set that appears valid while excluding the true pose. Such failures cannot be detected from the output, and they are precisely the failures that uncertainty quantification is meant to prevent. Calibration on translated images reduces them to a few percent of the test images, while most of the remaining failures are flagged by an empty pose set.

\notation{Conservative coverage.} Translated calibration restores valid coverage for $\alpha \geq 0.1$, but the coverage it achieves varies little with $\alpha$ and exceeds the specified level by a wide margin for larger values of $\alpha$. The guarantee is therefore recovered in a conservative rather than an exact form. This conservativeness has a cost: the keypoint disks are larger than necessary, and the average pose computed from them is less accurate at small values of $\alpha$. In practice, this suggests selecting a moderate $\alpha$, for which translated calibration still provides high coverage while the accuracy of the average pose remains comparable to that obtained with synthetic calibration. 


\notation{Absence of a formal guarantee.} Calibration on translated images restores exchangeability only approximately. As discussed in Section~\ref{sec:method}, the remaining loss of coverage is bounded by the discrepancy between the score distributions on translated and real images, which cannot be measured without labelled real images. The coverage obtained on SPEED+ is therefore an empirical result, and it does not constitute a guarantee for other targets, illumination conditions or translation models.

\notation{Availability of real images.} The translation models require unlabelled real images of the target domain. In our experiments, these images were drawn from the same domain as the test images, which represents a favourable setting. During a mission, representative images may be only partially available before operations, for instance from ground test facilities, and conditions that are absent from these images may again lead to a loss of coverage.

\notation{Scope of the evaluation.} The evaluation is limited to SPEED+, whose real images were acquired with a mockup in a laboratory rather than in orbit, and to a single keypoint network. In addition, the non-conformity score of~\cite{yang2023object} is driven by the worst keypoint of each image, which contributes to the size of the disks.

\section{Conclusion}
\label{sec:conclusion}

This paper studied conformal pose estimation for spacecraft under the sim2real gap. We applied the conformal keypoint detection and geometric uncertainty propagation to spacecraft pose estimation, and showed on SPEED+ that calibration on synthetic images leads to a substantial loss of coverage on real images, together with a large proportion of silent failures in which a non-empty pose set excludes the true pose. To address this issue, we calibrated on synthetic images translated to the appearance of real images, which preserves the synthetic pose labels and requires no labelled real images. With both CycleGAN and DUNIT, this calibration restored a coverage above the specified level for $\alpha \geq 0.1$ on both real domains and reduced silent failures to a few percent of the test images, at the cost of larger prediction sets.

Future work will aim to reduce the conservativeness of translated calibration, for instance through nonconformity scores that account for the differences between keypoints, and to combine it with domain adaptation in order to obtain a more accurate keypoint network and therefore tighter prediction sets. Further directions include the evaluation on other datasets such as SPARK and SPADES, the extension to image sequences, in which temporal correlation challenges the exchangeability assumption, and the evaluation of the worst-case error bounds of~\cite{yang2023object} under the sim2real gap.

\section{Acknowledgment}
\vspace{-0.1cm}
The work was funded by Luxembourg National Research Fund under the
project reference BRIDGES/2025/IS/19855391/U-ADAPT and Infinite Orbits. Experiments were performed on the Luxembourg national supercomputer MeluXina at LuxProvide. 

\section*{\refname}
\begingroup
\renewcommand{\section}[2]{}
\setlength{\itemsep}{0pt}\setlength{\parskip}{0pt}
\bibliographystyle{unsrt}
\bibliography{references}
\endgroup

\end{document}